\documentclass[10pt,twocolumn,letterpaper]{article}

\usepackage{iccv}
\usepackage{times}
\usepackage{epsfig}
\usepackage{graphicx}
\usepackage{amsmath}
\usepackage{amssymb}
\usepackage{multirow}

\usepackage[breaklinks=true,bookmarks=false]{hyperref}

\iccvfinalcopy % *** Uncomment this line for the final submission

\def\iccvPaperID{****} % *** Enter the ICCV Paper ID here

\ificcvfinal\fi

\begin{document}

%%%%%%%%% TITLE
\title{ Once a MAN: Towards Multi-Target Attack via Learning Multi-Target Adversarial Network Once}
\author{Jiangfan Han$^{1 *}$, Xiaoyi Dong$^{2}$\thanks{Equal contribution.}
, Ruimao Zhang$^{1}$, Dongdong Chen$^{2}$, Weiming Zhang$^{2}$, \\Nenghai Yu$^{2 \dagger}$, Ping Luo$^{3}$, Xiaogang Wang$^{1}$\thanks{Nenghai Yu and Xiaogang Wang are the corresponding authors.}\\
$^{1}$CUHK-SenseTime Joint Laboratory, The Chinese University of Hong Kong, \\ $^{2}$University of Science and Technology  of China, Key Laboratory of Electromagnetic \\Space Information, The Chinese Academy of Sciences, $^{3}$The University of Hong Kong\\
{\tt\small\{jiangfanhan@link., ruimao.zhang@, xgwang@ee.\}cuhk.edu.hk,}\\
{\tt\small\{dlight@mail., cd722522@mail., zhangwm@, ynh@\}ustc.edu.cn, pluo@cs.hku.hk }
}
%\and

\maketitle
\ificcvfinal\thispagestyle{empty}\fi

%%%%%%%%% ABSTRACT
\begin{abstract}
Modern deep neural networks are often vulnerable to adversarial samples. Based on the first optimization-based attacking method, many following methods are proposed to improve the attacking performance and speed. Recently, generation-based methods have received much attention since they directly use feed-forward networks to generate the adversarial samples, which avoid the time-consuming iterative attacking procedure in optimization-based and gradient-based methods. However, current generation-based methods are only able to attack one specific target (category) within one model, thus making them not applicable to real classification systems that often have hundreds/thousands of categories. In this paper, we propose the first Multi-target Adversarial Network (MAN), which can generate multi-target adversarial samples with a single model. By incorporating the specified category information into the intermediate features, it can attack any category of the target classification model during runtime. Experiments show that the proposed MAN can produce stronger attack results and also have better transferability than previous state-of-the-art methods in both multi-target attack task and single-target attack task. We further use the adversarial samples generated by our MAN to improve the robustness of the classification model. It can also achieve better classification accuracy than other methods when attacked by various methods.
\end{abstract}

%%%%%%%%% BODY TEXT
\vspace{-3mm}
\section{Introduction}
Deep neural networks (DNNs) have been significantly successful in many artificial intelligence tasks such as image classification \cite{krizhevsky2012imagenet,taigman2014deepface,he2016deep}, object detection \cite{girshick2015fast,ren2015faster,lin2017feature}, and natural language processing \cite{vaswani2017attention,gehring2017convolutional}. However, recent studies~\cite{szegedy2013intriguing,nguyen2015deep,moosavi2016deepfool} showed that DNNs are vulnerable and easy to be attacked by adversarial samples. By adding small perturbation onto an image, it is difficult to differentiate adversarial sample from the original image visually but can mislead modern classification model into definitely incorrect categories easily, which may produce severe security threat for real application systems such as auto-driving and face verification.
Adversarial samples also possess the characteristic of transferability, which means adversarial sample generated for attacking one model could also mislead another model.
Thus much attention has been concentrated on this phenomenon to better understand the weaknesses of DNNs, as well as to develop and strengthen the robustness of deep networks.

\begin{figure}[t]
\begin{center}
    \includegraphics[scale=0.65]{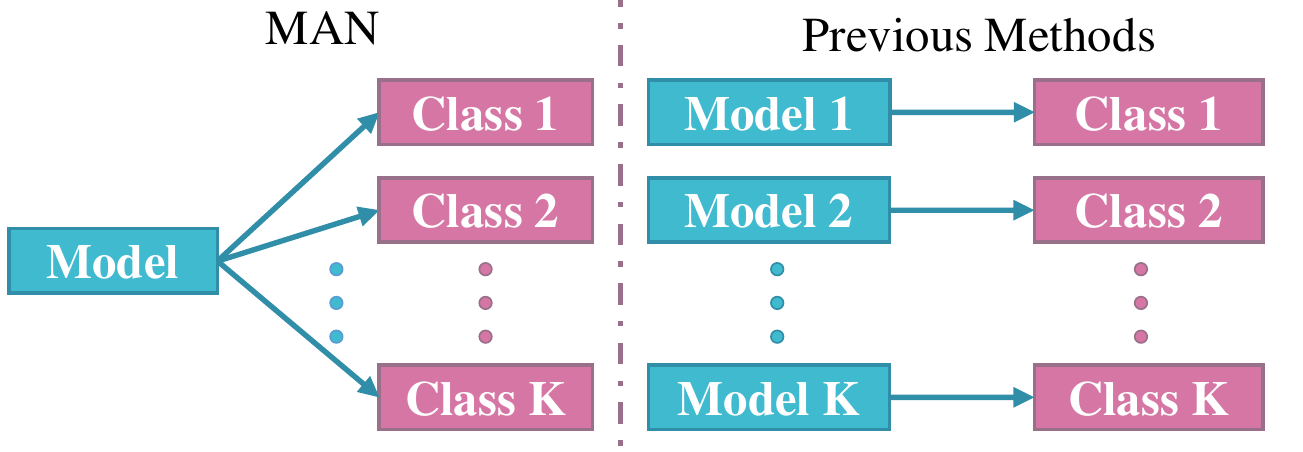}
\end{center}
   \caption{The comparison between our MAN and previous target attack methods. Our network can generate adversarial sample to all classes by training a single model once while others need individual models for each class.}
\label{fig:man_vs_single}
\end{figure}

The pioneering work from \cite{szegedy2013intriguing} used an optimization-based method to generate adversarial samples. Then many different types of methods \cite{nguyen2015deep,moosavi2016deepfool,papernot2017practical,su2019one} were proposed to improve the attacking performance and speed. For example, Goodfellow \etal \cite{43405} proposed the first gradient-based method, which can attack DNNs effectively by using the back-propagated gradient to update the input image with small perturbations iteratively. Moosavi-Dezfooli \etal \cite{moosavi2016deepfool} designed an iterative linearization of the classifier to generate minimal perturbations that are sufficient to change classification label. Although these optimization-based and gradient-based methods can produce very good attacking results, they often rely on a very time-consuming iterative procedure, which impose a great computation burden for real attacking systems. 

Recently, generation-based methods \cite{baluja2017adversarial,poursaeed2017generative,xiao2018generating} received much attention in the literature. They directly trained a generation model to learn how to transform input images to adversarial samples. This kind of methods can be viewed as the acceleration versions of above optimization-based and gradient-based methods. By training on a large scale of images with the pre-defined attacking object, these models do not need to access the target model again and can generate adversarial samples just with a fast forward pass during the runtime. However, one of the critical problems of current generation-based methods is that they are only able to realize single-target attack, meaning that adversarial samples generated by one model can only mislead the prediction of the attacked model to one specific predefined category during training. If we want to mislead the attacked model to another category, a new model needs to be trained, which is both time and storage consuming. Therefore, they are not feasible to real classification systems which contain hundreds/thousands of categories.

Motivated by these shortcomings, we propose a novel framework called {\em Multi-target Adversarial Network (MAN)}, which aims to generate multi-target attack samples by training the adversarial model once.
As illustrated in Figure~\ref{fig:man_vs_single}, MAN does not require specific attack target in training and can produce target adversarial samples for all categories in a dataset. 
Compared with existing single-target attack methods, MAN just trains once for all categories rather than training different models for different categories.

As shown in Figure~\ref{fig:overview}(a), MAN adopts the encoder-decoder network to embed the target information from incorrect categories and appearance information from input images to generate adversarial samples. 
It accomplishes the target-independent adversarial training by a simple framework with intuitive loss functions.
Moreover, through extensive experiments, we find that the adversarial sample generated by such a model has strong transferability. 
Figure~\ref{fig:vis_image} illustrates the adversarial samples generated by MAN. 
When we adopt different target information, MAN enables the attack samples to effectively inherit the appearance information from the original image, while cheating the pretrained model with high fooling rates.

Our \textbf{contributions} are summarized as follows.

\begin{itemize}
\item To the best of our knowledge, this is the first work to present the task of multi-target attack that generates target adversarial samples for all the categories in a dataset using one single model.
We accomplish this purpose by presenting a novel Multi-target Adversarial Network (MAN), which can produce multi-target adversarial samples by training a single model, and can significantly reduce the training cost and model storage.

\item In the single-target scenario, the proposed method achieves competitive and better attack performance and generalization ability, compared to state-of-the-art methods \cite{baluja2017adversarial,poursaeed2017generative} in various popular deep architectures~\cite{simonyan2014very,he2016deep}. For example, MAN achieves 98.55\% attack accuracy on the CIFAR10 dataset against VGG16 and 88.95\% when transferring to ResNet32, outperforming \cite{poursaeed2017generative} by 4.39\% and 12.49\% respectively. For the multi-target attack scenario, our model can maintain these advantages by training the multi-target adversarial network only once.

\item The generated attack samples by MAN effectively improve the robustness and adversarial defense ability of attacked models. Take the attacked model pretrained on the CIFAR10 as an example, when using the strongest attack samples generated by MAN to finetune ResNet32, the classification accuracy is improved from 10.58\% to 81.14\%, outperforming other counterparts by 17.74\%.

\end{itemize} 
%----------------------------------------------------------------------
\section{Related Work}
%As more and more deep neural networks have been used for real application systems, their robustness and security grow to be a great concern. Szegedy \etal \cite{szegedy2013intriguing} first discover that a high-performance deep neural network can be easily attacked by only adding some small perturbations onto the input image. The vulnerability of DNNs may attribute to their extreme non-linearity and insufficient regularization of their purely supervised learning strategies. Many following works have been proposed to attack different kinds of application systems, such as semantic segmentation and object detection \cite{xie2017adversarial}, speech recognition \cite{carlini2018audio}, and deep reinforcement learning \cite{lin2017tactics}.
\subsection{Adversarial Attack Methods}
 Current methods for generating adversarial samples can be divided into three categories:  optimization-based~\cite{szegedy2013intriguing,carlini2017towards}, gradient-based~\cite{43405,kurakin2016adversarial,dong2017boosting}, and generation-based~\cite{baluja2017adversarial,poursaeed2017generative}. Optimization-based methods model the generation of adversarial samples as an optimization problem and use optimizers like box-constrained L-BGFS~\cite{fletcher2013practical} or Adam~\cite{kingma2014adam} to solve it, which are powerful but quite slow. Goodfellow \etal~\cite{43405} proposed the first gradients-based method Fast Gradient Sign Method (FGSM) and Alexey \etal~\cite{kurakin2016adversarial} performed small step size iteratively to get better attacking performance (I-FGSM). The speed of gradient-based methods is linearly related to the number of parameters of target models and iteration numbers, and is often faster than optimization-based methods. However, due to their inherent model-specific mechanism, previous gradient-based methods have relatively weak transferability. Dong \etal~\cite{dong2017boosting} further introduced momentum into the iterative steps (MI-FGSM) to get better transferability.
%
%-------------------------------------------------------------
\begin{figure*}[t]
\begin{center}
\includegraphics[width= 17.5cm ,scale=0.54]{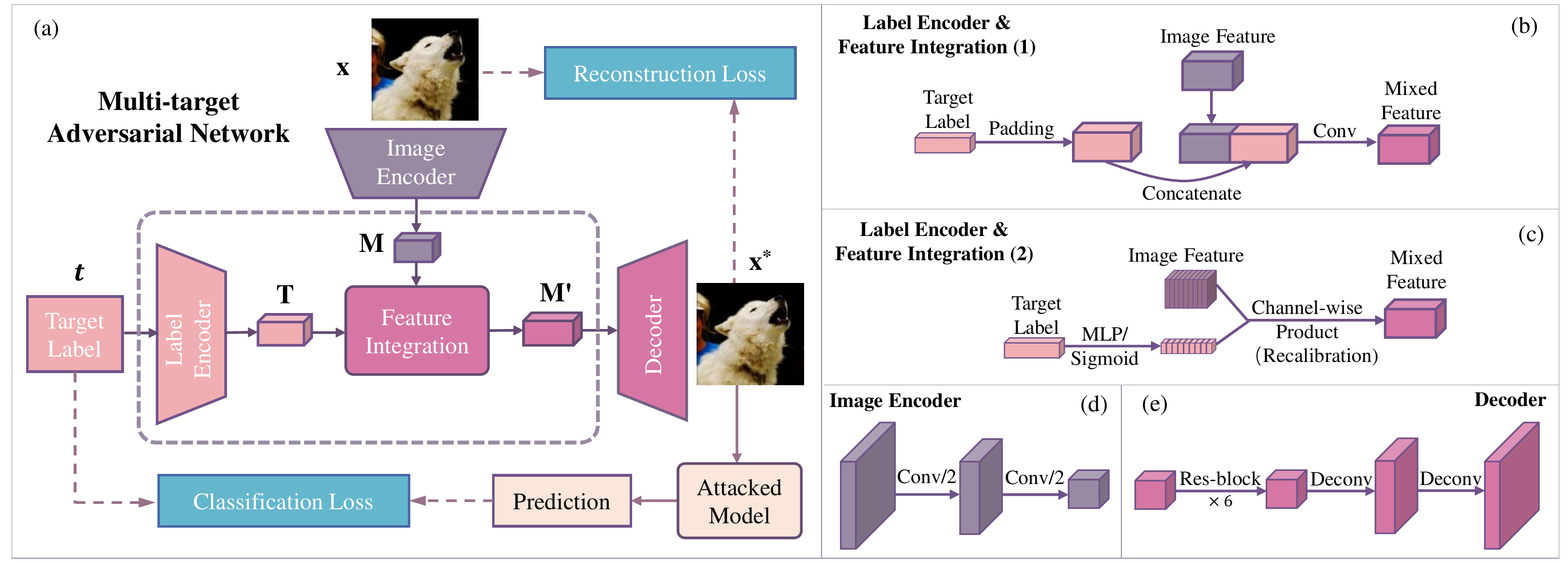}
\end{center}
\vspace{-3mm}
   \caption{The structure of Multi-Target Adversarial Network (MAN), (a) shows the overall architecture, (b) and (c) show two different label encoding and feature approaches, (d) and (e) are the architecture for image encoder and decoder respectively.}
\label{fig:overview}
\vspace{-3mm}
\end{figure*}
%-------------------------------------------------------------
Different from the iterative procedure used in the above two types of methods, generation-based methods directly use generation models to transform input images into adversarial samples. For example, an auto-encoder like network was adopted in ~\cite{baluja2017adversarial}, and U-Net and ResNet were leveraged in ~\cite{poursaeed2017generative}. Although these methods need extra time to train, they can generate adversarial samples at a constant speed and do not need to access to the target model again in the inference stage. The biggest limitation of all current generation-based methods is that they are only able to attack one specific category with one model, which is very unfriendly for real attacking systems. By contrast, the multi-target adversarial network proposed by us enables attacking any category of a classification model with similar attack success rate, and can also bring better transferability for the adversarial attack task as well as model robustness for the adversarial training task.

\subsection{Adversarial Defence Methods}
There were also many methods proposed to defend against adversarial attack. Adversarial training used in~\cite{2014arXiv1412.6572G, 2016arXiv161101236K, 2017arXiv170507204T} trained a robust network against adversarial attack by adding the adversarial sample into the training set and jointly trained with original samples.
In~\cite{2014arXiv1412.5068G, 7954632, 2017arXiv170502900D, 2017arXiv171202976L}, they used preprocessing procedures to remove the adversarial noise of the input images before feeding into the target model.
Other methods~\cite{2017arXiv170507204T,2016arXiv160202697P,2016arXiv161103814P} used some regularizes or smooth labels to make the target model more robust to the perturbation on input images. In the methods mentioned above, adversarial training is the most effective method. But it requires enough adversarial samples for training, so the speed of generating adversarial samples determines the effectiveness and performance of adversarial training. In this paper, we focus on how to generate adversarial sample to attack all categories with high speed, and this will be helpful for training more robust models.
% Current methods for adversarial defense can be divided into three categories: adversarial training based, preprocessing based and gradient masking based. Adversarial training based method\cite{2014arXiv1412.6572G, 2016arXiv161101236K, 2017arXiv170507204T} defense adversarial attack by training a robust model with adversarial samples. 
% Preprocessing based methods\cite{2014arXiv1412.5068G, 7954632, 726791, 2017arXiv170502900D, 2017arXiv171202976L} use some image operations to remove the adversarial noise of input images before feeding to the target model. Gradient masking based methods\cite{2017arXiv170507204T,2016arXiv160202697P,2016arXiv161103814P} use some regularizes or smooth labels to make the target model more robust to the perturbation on input images. In the methods mentioned above, adversarial training is the most effective method. But it requires enough adversarial samples for training, so the speed of generating adversarial samples determine the effectiveness and performance of adversarial training. In this paper, we focus on how to generate adversarial sample to attack all categories with high speed, this will be helpful for training more robust models.

\section{Multi-Target Attack}
%--------------------------------------------------------------
\textbf{Non-target \vs Single-target attack.} 
Let $\mathcal{H}$ be a deep neural network to be attacked, which is trained on a dataset with $K$ classes.
For an image $\mathbf{x}$ with the category label $y \in \{1,2,...,K\}$, 
the predicted probability of the network $\mathcal{H}$ for the $i$-th category is represented by $\mathcal{H}_i(\mathbf{x})$, where $i \in \{1,2,...,K\}$.
The non-target adversarial attack tries to generate a new image $\mathbf{x}^*$, 
such that $\mathrm{argmax}_{i}~\mathcal{H}_i(\mathbf{x}^*)\ne y$,
It implies that predicted label of $\mathbf{x}^*$ should not be the ground-truth label of $\mathbf{x}$.
By contrast, single-target attack aims to generate the adversarial sample $\mathbf{x}^*$,
such that $\mathrm{argmax}_{i}~\mathcal{H}_i(\mathbf{x}^*) = t$, 
where $t\ne y$ is a predefined class label that is specified in the testing stage.
According to the above definition, the non-target attack is much easier to realize than single-target attack, 
since it only requires the attacked model to make different predictions between the adversarial image and the corresponding original image.
Beyond the single-target attack, this work focuses on a more challenging task, multi-target attack, 
whose goal is to generate multi-target adversarial samples by a single network.

\textbf{White-Box attack \vs Black-Box attack.} 
If we have the knowledge about the attacked model $\mathcal{H}$, including model architecture and parameters, we call such attack as a white-box attack.
In this case, the attacker can generate adversarial samples directly using attack methods by accessing the attacked model.
On the contrary, for black-box scenario, the attacker knows nothing about the target model. Therefore the designed attacking model need stronger transferability, which means that adversarial samples generated for attacking one model could also mislead another model. 

\textbf{Problem Definition.}
In order to generate a target adversarial sample $\mathbf{x}^*$,
a transformation function $\mathcal{F}_\theta$ with parameters $\theta$ is defined to map the input image $\mathbf{x}$ to a target domain, 
${\bf{x}}^* = \mathcal{F}_\theta( \mathbf{x}, t )$.
In practice, we approximate this function by a deep neural network, where $\theta$ represents the network parameters.
% of the networks.
The purpose of multi-target attack is to train a network $\mathcal{F}_\theta$ 
such that $\mathrm{argmax}_{i}~\mathcal{H}_i(~\mathcal{F}_\theta( \mathbf{x}, t )~) = t$, where $t \ne y$.
Unlike traditional single-target attack that trains specific network $\mathcal{F}^{t_0}$ for a predefined label $t_0$, 
multi-target attack only trains a single network $\mathcal{F}$ for any target label $t \in \{1,2,...,K\}$.

%-------------------------------------------------------------------------
\subsection{Multi-Target Adversarial Network (MAN)}

\textbf{Architecture Overview.}
%
% The proposed Multi-Target Adversarial Network (MAN) is a natural extension of the single-target adversarial network.
%
The overall architecture of MAN is shown in Figure~\ref{fig:overview}~(a).
It takes image $\mathbf{x}$ and target label $t$ as input and generates adversarial sample $\mathbf{x}^*$ corresponding to the desired target.
Compared with traditional single-target adversarial networks~\cite{poursaeed2017generative},
the target label $t$ is regarded as a discrete variable rather than a constant in MAN. The range of $t$ is from $1$ to $K$.
Accordingly, the encoding network of MAN includes two branches, one for extracting the appearance features from the input image, as shown in Figure~\ref{fig:overview}~(d), 
and the other for encoding the target label information.
The feature integration module is also introduced to integrate the feature representation from these two modalities.
As shown in Figure~\ref{fig:overview}~(e), the decoder network with six residual blocks and two deconvolutional layers are adopted to generate the final adversarial samples.   

Two intuitive losses are employed to optimize the adversarial network. 
One is the reconstruction loss to preserve the appearance similarity between the original image and the adversarial samples
and the other is the classification loss which is adopted to embed the target label information into the attack samples.
To calculate the classification loss, the generated attack sample is fed into the attacked model, \ie the pretrained classification network, to predict the label.
Although the parameters in the attacked model are fixed during the MAN training, the gradients can still be passed through this model to guide the generative subnetwork.
%-------------------------------------------------------------------------

\textbf{Variants of MAN.}
Next, we describe two versions of feature integration schemes of MAN.
The first one is illustrated in Figure~\ref{fig:overview}~(b), 
which is a quite straightforward idea to integrate the image feature and label feature by concatenating along the channels.
Given an input image $\mathbf{x}$, the image encoder first calculates the image feature map  $\mathbf{M} \in \mathbb{R}^{C\times H \times W}$, where $C$, $H$ and $W$ indicate the number of channels, height and width of feature maps respectively.
The target label $t$ is presented as the one-hot vector $\mathbf{t} \in \mathbb{R}^{K}$,
where $\mathbf{t}$ is expanded along height and width directions to get the label feature map 
$\mathbf{T} \in \mathbb{R}^{K \times H \times W}$.
Then the above two sets of feature maps are concatenated along the channels to get a mixed one ${\tilde{\mathbf{M}}}' \in \mathbb{R}^{(K+C) \times H \times W}$.
An additional convolutional layer is adopted to reshape the mixed feature ${\tilde{\mathbf{M}}}'$ to $\mathbf{M}' \in \mathbb{R}^{C \times H \times W}$. 
At last, $\mathbf{M}'$ is fed into the decoder to generate the adversarial sample ${\bf{x}}^*$.
In this approach, the mixed feature is generated by concatenating the feature maps of image and target label.
We denote this model as the multi-target adversarial concatenate network (MANc).

%------------------------------------------------------------------------

The second architecture of label encoder and feature integration is illustrated in Figure~\ref{fig:overview}~(c). 
Inspired by the Squeeze and Excitation Network proposed by Jie \etal~\cite{hu2017squeeze}, 
we claim that different channels of feature representation can capture diverse features for different classes.
Thus the channel-wise product, \ie recalibration operation, is used to integrate the label feature and image feature.
In such a case, the label encoder contains a two-layer multi-layer perceptron (MLP),
and the output is activated by sigmoid function $\sigma(x) = \frac{1}{1+e^{-x}}$.
We denote $\mathbf{t}'$ as the output of MLP, which is given by:
\begin{align}
    {\mathbf{t}'} = {\mathcal{F}}_{en}({\bf{W}} , {\bf {t}}) = \sigma({\bf{W}}_2\delta({\bf{W}}_1 {\bf {t}})),
\end{align}
where $\delta$ is the ReLU~\cite{nair2010rectified} activation function, ${\bf{W}}_1 \in \mathbb{R}^{U \times K}$ and ${\bf{W}}_2 \in \mathbb{R}^{C \times U}$ are fully connected layers, $U$ is the number of intermediate units, $\mathbf{t}' \in \mathbb{R}^{C}$, and $\sigma$ limits every element $t'_i \in (0,1)$.
The final integrated feature representation is presented as $\mathbf{M}' = \mathbf{t}' \odot \mathbf{M}$,
where $\odot$ indicates the channel-wise product on the tensor: $\mathbf{M}'_{i} = t_i \cdot {\mathbf{M}}_{i}$, in which ${\bf{M}}_{i} \in \mathbb{R}^{H \times W}$ is channel-wise feature map of image feature ${\bf{M}} = [{\bf{M}}_1 , {\bf{M}}_2, \dots , {\bf{M}}_C]$.
We refer this structure as multi-target adversarial recalibrating network (MANr).
%--------------------------------------------------------------
\subsection{Optimization}

%--------------------------------------------------------------
\begin{figure}[t]
\begin{center}
\includegraphics[scale=0.67]{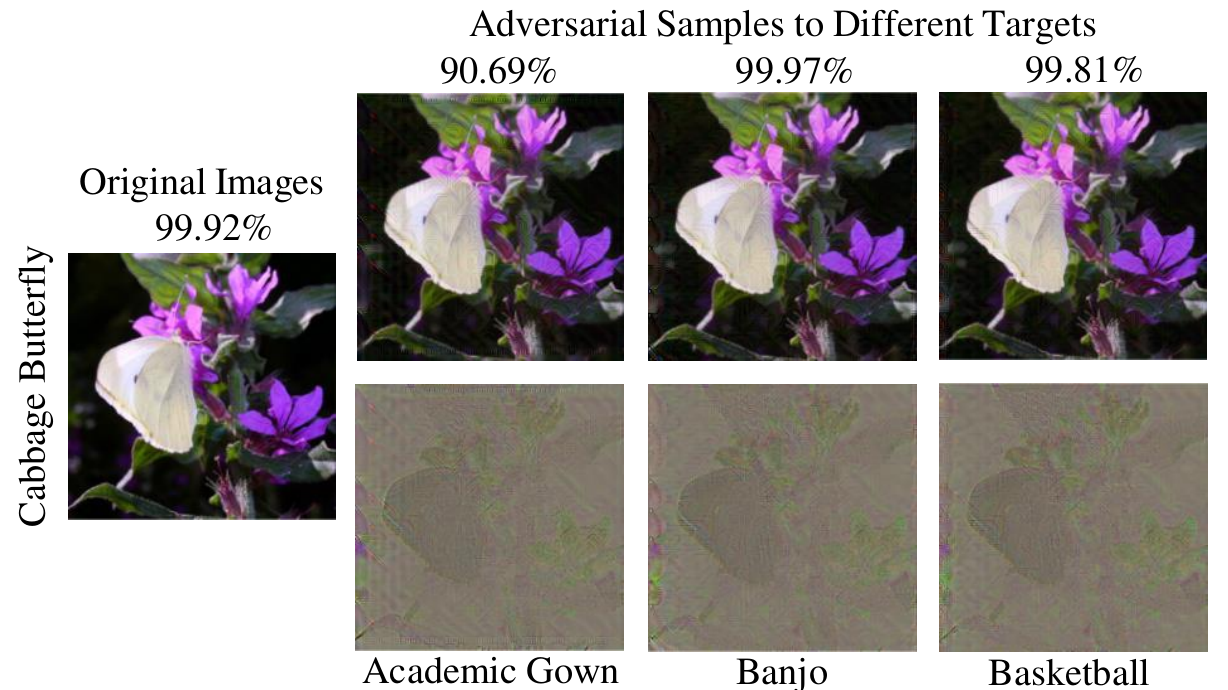}
\end{center}
\vspace{-3mm}
   %\caption{Adversarial samples to different target with $\epsilon=10\sqrt{N}$ compare with the original images. \textbf{Left:} the original images with ground truth label selected from ImageNet. \textbf{Right:} the adversarial samples to different targets (lower columns show the difference between adversarial samples and original images). Numbers are the predicted probability for the corresponding label given by a pretrained VGG16 model. }
   \caption{Adversarial samples to different target with $\epsilon=10\sqrt{N}$. Numbers are the predicted probability for the corresponding label given by a pretrained VGG16 model. }
\label{fig:vis_image}
\vspace{-5mm}
\end{figure}
%---------------------------------------------------------------
%
To optimize the network parameters, we define the objective function of MAN for target label $t$ as follows,
\begin{align}
    \mathcal{L} (\mathbf{x}) = \mathcal{L}_{cls}(~\mathcal{H} (  \mathcal{F}_\theta (\mathbf{x}, t )  ), t~) + \alpha \mathcal{L}_{re}(~\mathbf{x} , \mathcal{F}_\theta (\mathbf{x}, t)),
 \label{loss-multi}
\end{align}
where $\mathcal{H} ( \mathcal{F}_\theta  (\mathbf{x},t))$ is the predicted probability of the target model on adversarial sample,
$t$ is the discrete label variable, weight factor $\alpha$ decides the importance of two loss terms,
and $\mathcal{L}_{cls}$ is the classification loss which encourages the attacked model make prediction to the target label.
The concrete format of loss function has massive impact on the final result.
Previous studies~\cite{baluja2017adversarial,poursaeed2017generative, carlini2017towards} tried different classification losses in their works;
we found that the standard cross-entropy loss works well in all of our experiments.

The reconstruction loss $\mathcal{L}_{re}$ measures the appearance difference between  adversarial sample and original image.
It is usually measured by vector norm $|| \mathbf{x} - \mathcal{F}_\theta (\mathbf{x},t) ||_p$ where $p\in\{0,1,2,\infty\}$~\cite{carlini2017towards}.
In this paper, we adopt Euclidean distance, \ie $L_2$ norm, as our measurement.
To ensure the difference between adversarial sample and the original image is small, \ie the perturbation is restricted to a range.
A fixed distance upper bound $\epsilon$ is given and we force the constraint $||  \mathbf{x} - \mathcal{F}_\theta (\mathbf{x},t) ||_2 < \epsilon$.
% 
%5In practice, we consider two different approaches to achieve this constraint. 
In practice, we use $\alpha$ to leverage the power of the two loss terms.
Larger $\alpha$ encourages better reconstruction quality while smaller ones get higher success attack rate.
%
%5In the first one, we scale the norm of perturbation to the

%---------------------------------------------------------------
\section{Experiments }
%---------------------------------------------------------------
\subsection{Adversarial Attack}
\label{setup}
In this section, we evaluate the attack effectiveness of MAN on the ImageNet~\cite{deng2009imagenet} and the CIFAR10~\cite{krizhevsky2009learning}.

\textbf{General setup}.
On the ImageNet~\cite{deng2009imagenet} dataset, we conduct the experiments by using popular network architectures VGG16~\cite{simonyan2014very}, and ResNet152 (Res152)~\cite{he2016deep} as attacked models . 
Meanwhile, we use VGG19 and Res101 as ``black-box'' model.
For fair comparison, in the testing stage, we scale the norm of all perturbations to a certain threshold $\epsilon$.
\ie $\mathbf{\hat{x}^*} = \mathbf{x} - \frac{\epsilon }{|| {\bf{x}} - \mathcal{F}_\theta(\mathbf{x},t) ||_2} ( {\bf{x}} - \mathcal{F}_\theta(\mathbf{x},t))$. 
Here $\mathbf{\hat{x}^*}$ is the scaled adversarial sample for testing.
The threshold of the $L_2$ norm $\epsilon$ in our experiments is calculated by $\delta \sqrt{N}$, where $N = C \times H \times W$ is the dimension of input images $\mathbf{x}$.
%Here we set different $\delta = 6, 8 ,10$.
We set $\delta = 10$ in this part and further exploration on $\delta$ is listed in the ablation study.
%
% while the pixel value in each image is in the range of [0,255]. 
%
We use Adam \cite{kingma2014adam} optimizer with $\beta_1 = 0.5$ and $\beta_2 = 0.999$ to train all of the adversarial networks.
The batch size is 32.
%
%the hyperparameters of Adam optimizer are the same as the experiments in ImageNet and the batch size is 128.
For the CIFAR10 \cite{krizhevsky2009learning}, the  attacked networks are VGG16~\cite{simonyan2014very} and ResNet32 (Res32)~\cite{he2016deep}, and ``black-box'' models are VGG19 and Res14. We set batchsize to 128, and other settings are the same as the Imagenet.

%Using a similar approach, the result on the ImageNet also split into two parts. The first part considering the single-target case and the second part is focused on the multi-target task.

%-------------------------------------------------------------------------
%single-image
%-------------------------------------------------------------------------
\begin{table}
\setlength{\tabcolsep}{1.3mm}
\small
\begin{center}
\begin{tabular}{c|c|c|c|c|c}
\hline
\multicolumn{2}{c|}{Attacked model} &VGG16& VGG19& Res152 & Res101\\
\hline\hline
\multirow{4}*{VGG16} &
ATN~\cite{baluja2017adversarial}& 88.18$^*$ & 55.68 & 0.15  &  0.13  \\
&GAP~\cite{poursaeed2017generative}& 99.89$^*$& 70.83& 0.53 & 0.30 \\
&MANc & 99.85$^*$& \bf{85.79}& \bf{0.90} & \bf{0.50}\\
&MANr &\bf{99.93$^*$} & 71.83 & 0.50 & \bf{0.50}  \\
\hline
\multirow{4}*{Res152} &
ATN~\cite{baluja2017adversarial}& 0.07 & 0.08 & 81.95$^*$ & 2.31 \\
&GAP~\cite{poursaeed2017generative}& 19.17& 16.51 & 99.65$^*$ & 73.53 \\
&MANc & 3.58 & 5.48 & \bf{99.67$^*$} & 52.75 \\
&MANr & \bf{20.47} & \bf{23.74} & 99.72$^*$ & \bf{77.61}  \\
\hline

\end{tabular}
\end{center}
\caption{The success rate (\%) of single-target adversarial attack on the ImageNet dataset with different target model. $^*$ indicates the white-box attacks.}
\label{single-image}
\end{table}
%-------------------------------------------------------------------------

%-------------------------------------------------------------------------
%single-cifar
%-------------------------------------------------------------------------
\begin{table}
\setlength{\tabcolsep}{1.3mm}
\small
\begin{center}
\begin{tabular}{c|c|c|c|c|c}
\hline
\multicolumn{2}{c|}{Attacked model} &VGG16& VGG19& Res32 & Res14\\
\hline\hline
\multirow{4}*{VGG16} &
ATN~\cite{baluja2017adversarial}& 87.97$^*$ & 65.71 & 69.85 & 71.81 \\
&GAP~\cite{poursaeed2017generative}& 94.16$^*$ & 77.60 & 76.46 & 78.90 \\
&MANc & 97.86$^*$ & 87.67 & 88.26 & 88.63 \\
&MANr & \bf{98.55$^*$} & \bf{91.07} & \bf{88.95} & \bf{90.33}  \\
\hline
\multirow{4}*{Res32} &
ATN~\cite{baluja2017adversarial}& 37.41 & 50.74 & 93.17$^*$ & 69.60 \\
&GAP~\cite{poursaeed2017generative}& 43.71 & 57.19 & 97.02$^*$ & 74.30 \\
&MANc & 51.27 & 68.11 & 98.86$^*$ & \bf{81.01} \\
&MANr & \bf{58.56} & \bf{72.26} & \bf{99.26$^*$} & 79.33 \\
\hline

\end{tabular}
\end{center}
\caption{The success rate (\%) of single-target adversarial attack on the CIFAR10 dataset with different target model. $\ast$ indicates the white-box attacks. }
\label{single-cifar}
\end{table}
%-------------------------------------------------------------------------

\textbf{Single-target attack.}
\label{Single-exp}
In this part, we evaluate the attack performance on the single-target attack task.
We fix the input label $t$ during the single-target training process to get a degraded version of our model.
We compare the proposed MANc and MANr with two other state-of-the-art methods ATN~\cite{baluja2017adversarial} and GAP~\cite{poursaeed2017generative}.
% The former is a generation-based method and the later is a gradient-based one.
%The adversarial samples are generated respectively by GAP\cite{poursaeed2017generative}, MI-FGSM\cite{dong2017boosting} and MANc, MANr proposed by us.
%
For all of the methods, 
we train adversarial models for 120K iterations, with an initial learning rate of 0.002 and decreased by 10 after 80K iterations on the ImageNet. On the CIFAR10 training set, we train 20K iterations with an initial learning rate of 0.001 and decreased by 10 after 16K iterations. The hyperparameter $\alpha$ is set to 100 (1 for ATN) on the ImageNet and 800 (80 for ATN) on the CIFAR10.
During testing, adversarial samples are generated by 50K images in the ImageNet validation set and 5k images in the CIFAR10 validation set. We use the success rate as the evaluation metric, which is defined as the ratio of the attacked model classifying the adversarial samples to the target class. 

Table~\ref{single-image} and Table~\ref{single-cifar} report the attack success rates on two datasets.
We first evaluate the white-box attack performance. From the results, we can find that ATN performs quite weak for all cases. 
% on both ImageNet and CIFAR10.
%
Comparing with GAP, both MANc and MANr achieve better results in most cases, 
\eg  MANr achieves 98.55\% attack accuracy against VGG16 on the CIFAR10, outperforming GAP by 4.39\%.

When it comes to the black-box attack.
ATN shows very weak transferability. 
When transferring to a model with similar architecture, \eg from VGG16 to VGG19, our MANc get 85\% success rate on the ImageNet and nearly 90\% on the CIFAR10, outperform GAP by over 10\%.

When transferring to a model with totally different architecture \eg from VGG to ResNet, we find results on the CIFAR10 decrease a bit but still higher than 88\% in most cases. 
But on the ImageNet, the success rate of all methods are lower than 25\%.
We think this is because there are too many categories in the ImageNet dataset, making it hard to transfer to another model with a specific attack target. 

According to the above results, our method performs much better in most cases.
We think this is because our method splits the task into two branches. 
One branch guides the generation model to generate adversarial samples similar to input as the previous methods did.
The other branch encodes the target label and provides extra guidance helping generate more powerful adversarial samples.

%--------------------------------------------------------------
\begin{table}
\setlength{\tabcolsep}{1.3mm}
\small
\begin{center}
\begin{tabular}{c|c|c|c|c|c}
\hline
\multicolumn{2}{c|}{Attacked model} &VGG16& VGG19 & Res152 & Res101\\
\hline\hline
\multirow{2}*{VGG16} &
MANc & \bf{99.22$^*$}& \bf{66.81} &\bf{2.48}  & \bf{1.42} \\
&MANr & 99.13$^*$ & 54.16 & 1.73 & 1.33 \\
\hline
\multirow{2}*{Res152} &
MANc & \bf{7.61} & \bf{7.71} & 98.14$^*$ & \bf{55.39}\\
&MANr & 5.76 & 6.06 & \bf{98.23$^*$} & 49.69\\
\hline
\end{tabular}
\end{center}
 \vspace{-1mm}
\caption{The success rate (\%) of multi-target adversarial attack on the ImageNet dataset with different model. $\ast$ indicates the white-box attacks. }
\label{multi-image}
\end{table}
%---------------------------------------------------------------
\begin{table}
\setlength{\tabcolsep}{1.3mm}
\small
\begin{center}
\begin{tabular}{c|c|c|c|c|c}
\hline
\multicolumn{2}{c|}{Attacked model} &VGG16& VGG19 & Res32 & Res14\\
\hline\hline
\multirow{2}*{VGG16} &
MANc & 99.14$^*$& 92.97& 90.08 & 88.56\\
&MANr & \bf{99.50$^*$} & \bf{95.47} & \bf{92.68} & \bf{89.74} \\
\hline
\multirow{2}*{Res32} &
MANc & 76.86 & 86.87 & \bf{99.30$^*$} & \bf{90.27}\\
&MANr & \bf{78.36} & \bf{88.15} & 98.94$^*$ & 84.98\\
\hline
\end{tabular}
\end{center}
 \vspace{-1mm}
\caption{The success rate (\%) of multi-target adversarial attack on the CIFAR10 dataset with different model. $\ast$ indicates the white-box attacks. }
\label{multi-cifar10}
\vspace{-2mm}
\end{table}
%-------------------------------------------------------------------------
\textbf{Multi-target attack.}
In the training phase for multi-target attack, a random target label is assigned to each training image. We train models for 280K iterations, with an initial learning rate of 0.002 and decreased by 10 after 240K iterations on the ImageNet training set. On the CIFAR10 training set, we train 80K iterations, with an initial learning rate of 0.001 and decreased by 10 after 64K iterations.
At the evaluation phase, we randomly pick 5000 samples from the ImageNet validation set and randomly assign ten labels for each sample. For the CIFAR10, the validation set contains 5000 samples and we assign all of the ten labels to each sample. In total, there are 50K adversarial samples generated to attack the pretrained models on each dataset. 

Table \ref{multi-image} and Table \ref{multi-cifar10} report the attack success rates of variants of MAN on the ImageNet and on the CIFAR10 respectively. 
For white-box case, we find our models keep a great performance with attack accuracy over 98\%.
%and achieve comparable success rates comparing to the single target case. \textbf{}
When evaluating the black-box performance on both datasets, we find our models also keep a high transferability.
Since the evaluation metric in Table~\ref{single-image},~\ref{single-cifar} and Table~\ref{multi-image},~\ref{multi-cifar10} are different, the result can not be compared directly. We make further comparison under the same evaluation metric in the ablation study.

Results show the feasibility and effectiveness of our methods to realize the multi-target attack task. Even on a large dataset with numerous categories, our method can still attack arbitrary category with a single model.

There is also an interesting phenomenon that MANr performs better on CIFAR10 while MANc has better results on ImageNet. MANr re-calibrates each channel of the feature map. It controls better when size is small. When size is large, it is not easy to re-calibrate the feature map. MANc concatenates feature maps to get a mixed one, which is not influenced by the size of feature map and thus performs evenly on both datasets.
Figure~\ref{fig:vis_image} shows the original samples and adversarial samples generated by different target label.
%---------------------------------------------------------------
\begin{table}
\setlength{\tabcolsep}{1.3mm}
\small
\begin{center}
\begin{tabular}{c|c|c|c}
\hline
Method & No. of Params & Total Models  & Total Training Iters\\
\hline
\hline
ATN~\cite{baluja2017adversarial}&7.84M &1000 &120M \\
GAP~\cite{poursaeed2017generative}&7.84M &1000 &120M \\
MANc&10.74M&1&280K \\
MANr&8.55M&1&280K\\
\hline
\end{tabular}
\end{center}
 \vspace{-3mm}
\caption{Comparison of our method with single-target methods on Number of parameters for each model, the total number of models used  and total training iteration needed to realize attack all categories in the ImageNet.}
\vspace{-4mm}
\label{imagenet-compare}
\end{table}

\textbf{Parameter Comparison}.
Table~\ref{imagenet-compare} lists the number of parameters for each model, the number of models needed and the total training iterations needed to attack all categories in the ImageNet.
Previous models have about 8.3\% fewer parameters comparing with MANr, but 1K models with in total 7840M parameters and 120M iterations are required to attack all categories in the  ImageNet.
By contrast, our method just adopts a single model with much fewer parameters and 0.3M  iterations to achieve the same goal.
Thus our method is significantly more time and storage efficient.

%---------------------------------------------------------------
\begin{table*}
\begin{center}
\begin{tabular}{c|c|c|c|c|c|c|c}
\hline
Attack Strength& Attack Method&GAP 1 & GAP 5 & GAP 10 & MANc & MANr & Raw Res32\\
\hline\hline
\multirow{2}{*}{$8\sqrt{N}$} &ATN~\cite{baluja2017adversarial} &13.38 & 11.60 &  11.37 & \bf{11.17} &11.43 & 86.41 \\
&MI-FGSM~\cite{dong2017boosting} & 18.18 & 13.96 & 13.76 & \bf{12.34} & 13.24 & 99.40 \\
\hline
\multirow{2}{*}{$12\sqrt{N}$} &ATN~\cite{baluja2017adversarial} &19.40 & 13.50 & 13.08 & \bf{12.57} &13.28 & 94.56 \\
&MI-FGSM~\cite{dong2017boosting} & 23.56 & 18.86 & 15.28 & 14.86 & \bf{13.84} & 99.68 \\
\hline
\multirow{2}{*}{$16\sqrt{N}$} &ATN~\cite{baluja2017adversarial}&24.93 & 16.09 & 15.27 & \bf{14.69} & 15.88 & 97.97 \\
&MI-FGSM~\cite{dong2017boosting} & 29.22 & 22.22 & 17.44 & 17.18 & \bf{16.88} & 99.76 \\
\hline
\end{tabular}
\end{center}
\vspace{-2mm}
\caption{Attack success rate (\%) against classification models. Lower success rate indicate the model is more robust. ResNet32 is the attacked model to generate adversarial samples by MI-FGSM and ATN with $L_2$ threshold $8\sqrt{N}$,$12\sqrt{N}$,$16\sqrt{N}$. We also add Raw Res32 (ResNet32 without adversarial training) to make sure the effectiveness of attack methods.}
\label{adv_attack}
\end{table*}
%-------------------------------------------------------------------------
\begin{table*}
\begin{center}
\begin{tabular}{c|c|c|c|c|c|c|c}
\hline
Attack Strength& Attack Method&GAP 1 & GAP 5 & GAP 10 & MANc & MANr & Raw Res32\\
\hline\hline
\multirow{2}{*}{$8\sqrt{N}$} &ATN~\cite{baluja2017adversarial} &86.17 & 89.48& 89.85 & \bf{89.91} & 89.36 & 20.77 \\
&MI-FGSM~\cite{dong2017boosting} & 81.90 & 86.60 & 88.74 & 89.24 & \bf{89.34} & 9.54 \\
\hline
\multirow{2}{*}{$12\sqrt{N}$} &ATN~\cite{baluja2017adversarial} &75.48 & 85.38 & 85.70 & \bf{86.06} & 84.87 & 13.04 \\
&MI-FGSM\cite{dong2017boosting} & 73.70 & 81.40 & 84.78 & \bf{86.14} & 86.06 & 10.12 \\
\hline
\multirow{2}{*}{$16\sqrt{N}$} &ATN~\cite{baluja2017adversarial} & 63.41 & 78.87 & 78.64 & \bf{79.15} & 77.17 & 10.78 \\
&MI-FGSM~\cite{dong2017boosting} & 63.40 & 73.86 & 79.08 & \bf{81.14} & 81.00 & 10.58 \\
\hline
\end{tabular}
\end{center}
\vspace{-2mm}
\caption{Classification accuracy (\%) for classification models. Higher accuracy indicate the model is more robust.}
\label{adv_accu}
\vspace{-4mm}
\end{table*}
%---------------------------------------------------------------
%
\subsection{Adversarial Training}
\label{sub:adv_train}
Adversarial training is one of the most effective method to improve the robustness of models against adversarial attack.
It finetunes the network by adversarial samples with the groundtruth labels of the images used to generate the adversarial samples.
In this part, we evaluate the improvement of classification model against adversarial attack when the adversarial samples are used for finetuning.

\textbf{Setups}.
All of the following experiments are conducted on the CIFAR10 dataset.
To improve the robustness of the classification model, we use adversarial samples and their ground truth labels to finetune the pretrained networks.
In the rest part of the paper, we use ``adv-model'' to indicate the model finetuned by the adversarial samples.
The attacked model architecture is ResNet32, and the adversarial samples are generated by MANc, MANr and GAP~\cite{poursaeed2017generative} respectively with $\epsilon = 10\sqrt{N}$.

For MANc and MANr, we generate adversarial samples with random target labels and finetune the attacked model with these samples.
We also train ten GAP models with respect to ten attack target, each of which is a single-target attack model, to produce adversarial samples.
We finetune the attacked model with adversarial samples generated by one, five and ten GAP models to explore the influence of the number of classes used during adversarial training. 
The corresponding settings are denoted as GAP1, GAP5, and GAP10 respectively.
All the adv-models are finetuned with 40K iterations with batch size 128 (64$\times$2, each image is coupled with its adversarial sample). The learning rate is set to 0.001.
To further evaluate the robustness of these adv-models, we use MI-FGSM~\cite{dong2017boosting} and ATN~\cite{baluja2017adversarial} as attack methods.
In practice, we use MI-FGSM to generate adversarial samples with random target labels and follow the optimal settings used in ~\cite{dong2017boosting}. 
For ATN method, we follow the generation procedure stated in Sec.~\ref{Single-exp}.
Note that all the attack samples for testing are generated by using pretrained ResNet32 architecture on the CIFAR10 test set.

We consider the robustness of a classification model in two aspects.
On the one hand, we define the attack success robustness by the attack success rate of adversarial samples to the adv-models.
%
% We use attack success rate of adversarial samples to evaluate this kind of robustness.
%
In such a case, a lower success rate indicates the model is more robust. 
On the other hand, the classification success robustness is also defined to evaluate whether the adv-models could classify the adversarial samples to their ground truth classes.
In this subsection, we set the attack strength (threshold) $\epsilon = 8\sqrt{N}, 12\sqrt{N}, 16\sqrt{N}$. 
Two of these thresholds are larger than that in the last part because the adv-models has already finetuned by adversarial samples. 
A larger threshold can also reflect the robustness of adv-models against stronger attack.

\textbf{Attack success robustness.}
We show the result of attack success rates in Table \ref{adv_attack}.
According to the results, the attack success rate grows as the attack strength grows.
When the attack strength is small, \ie $8\sqrt{N}$, 
it is quite hard to use adversarial samples to fool any of the adv-models,
while GAP1 is slightly worse than others.
When the strength becomes larger, \ie $12\sqrt{N}$, GAP1 and GAP5 models tend to be vulnerable to adversarial samples, which almost increase the attack success rate by around 5\%, whereas GAP10 and our models are still robust, and the success attack rate is slightly increased by around 2\%.
When the attack strength is set to $16\sqrt{N}$, 
GAP1 and GAP5 let over 20\%  adversarial samples attack successfully.
Other models also tend to be vulnerable in such case, but our MANs still slightly outperform the GAP10. \eg MANr can defense 0.56\% more adversarial samples against MI-FGSM attack.
%----------------------------------------------------------------

\textbf{Classification success robustness.}
We report the results of classification success robustness in Table \ref{adv_accu}.
By using small attack strength, \ie $8\sqrt{N}$, all of the adv-models can classify most adversarial samples correctly and achieve high classification accuracy (larger than 80\%).
When the attack strength becomes larger, \ie $12\sqrt{N}$, the accuracy of the GAP1 model decrease obviously, while other adv-models can still maintain a promising accuracy over 80\%.
When setting the attack strength to $16\sqrt{N}$, only GAP10, MANc and MANr can keep the accuracy around 80\%. Moreover, our methods get better results. \eg For MI-FGSM attack method, MANc can get 81.14\% accuracy, outperforming GAP10 by 2.06\%

In summary, when we compare the adversarial training results among GAP1, GAP5 and GAP10 models, 
we find that when the number of target models increases, the adv-models can obtain more ability to resist the adversarial attack, which demonstrates the necessity to retrain the attacked model using multiple single-target attack models.
At the same time, compared with single-target methods, our models show better performance for almost all cases.
It demonstrates that our methods can generate diverse adversarial samples to promote the retraining efficiency of adv-models.
Our methods can use fewer training time and storage resources to produce richer adversarial samples and make the attacked models more robust.

%-----------------------------------------------------------------------
\subsection{Ablation Study}
\label{ablation}

\textbf{Weight factor $\alpha$}.
%----------------------------------------------
In this part, we explore the influence of weight factor $\alpha$ for attack result.
As we stated above, a larger $\alpha$ encourages better reconstruction quality while smaller ones lead to higher attack rate.
When evaluating the results, we compare the attack rate under a certain reconstruction threshold $\epsilon$, and the influence of it becomes complicated.

\begin{figure}[t]
    \centering
    \small
    \begin{tabular}{c@{\hspace{-6mm}}c@{\hspace{-3mm}}c}
        &
        \begin{tabular}{c}
           \includegraphics[width=4.3cm]{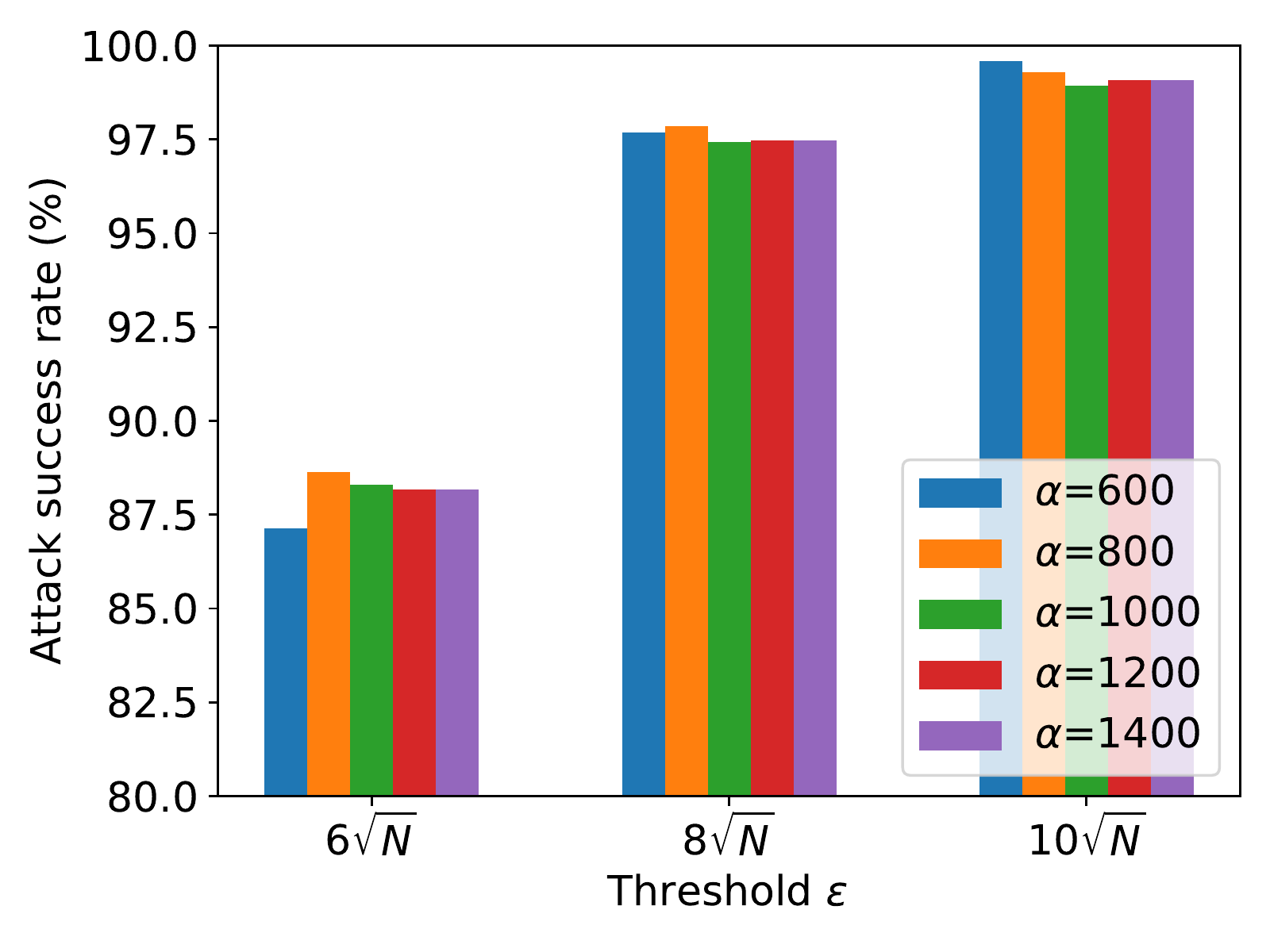}
        \end{tabular}
        &
        \begin{tabular}{c}
            \includegraphics[width=4.2cm]{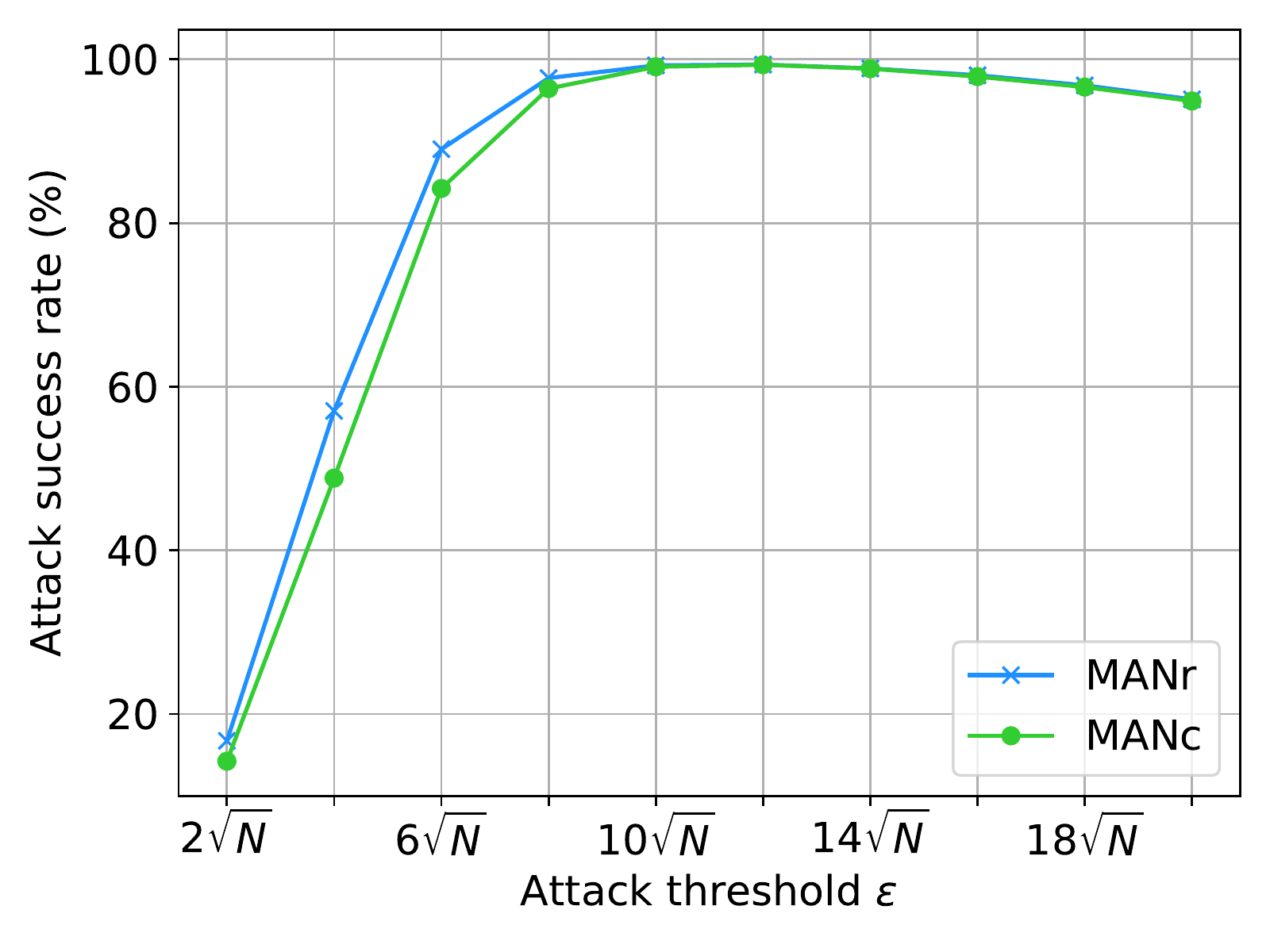}
        \end{tabular}\\
        & (a) & (b)\
    \end{tabular}
\caption{(a) The relation between weight factor $\alpha$ and attack success rate with different $\epsilon$ on the CIFAR10 to a pretrained VGG16 model. (b) The attack accuracy under different threshold $\epsilon$.}
\label{fig:alpha_image}
\vspace{-1mm}
\end{figure}

Figure~\ref{fig:alpha_image}~(a) shows the attack success rate on the CIFAR10 dataset with different weight factor $\alpha$ to a VGG16 model.
We can find that smaller $\alpha$ performs well when the permitted threshold $\epsilon$ is large.
Larger $\alpha$ often performs better when the $\epsilon$ at a small value, but sometimes it may depress the performance among all $\epsilon$ cases.

\textbf{Threshold $\epsilon$}.
In this part, we explore the influence of different threshold $\epsilon$.
%
% $\epsilon$ is the norm of perturbations. 
%
Larger $\epsilon$ means more severe change to the original image. 
We test the attack accuracy under different $\epsilon$ from $2\sqrt{N}$ to $20\sqrt{N}$ with step size $2\sqrt{N}$. 

Figure~\ref{fig:alpha_image}~(b) shows the attack success rate on the CIFAR10 dataset. 
We find that smaller $\epsilon$ (such as $2\sqrt{N}$, $4\sqrt{N}$) limits the success rate.
When $\epsilon$ increases, success rate grows at the same time. 
$10\sqrt{N}$ and $12\sqrt{N}$ get the best performance so we use $10\sqrt{N}$ in our main experiments. 
But when the $\epsilon$ is too large, the performance decreases a bit. 
We think this is because perturbation with too large $\epsilon$ impedes the representation of adversarial features.

%--------------------------------------------------------------
 \textbf{Transferability}.
 %--------------------------------------------------------------
% %------------------------------------------------------------------------
% \begin{figure}[t]
% \begin{center}
% \includegraphics[width=8.0cm ,scale=0.54]{per.png}
% \end{center}
%  \vspace{-3mm}
% \caption{(a) The relation between weight factor $\alpha$ and attack success rate with different $\epsilon$ on the CIFAR10 to a pretrained VGG16 model. (b) The attack accuracy under different threshold $\epsilon$.}
% \label{fig:alpha_image}
% \vspace{-1mm}
% \end{figure}
 %--------------------------------------------
 \begin{table}
 \setlength{\tabcolsep}{1mm}
 \small
 \begin{center}
 \begin{tabular}{c|c|c|c|c|c}
 \hline
  & & VGG16 & VGG19 & Res32& Res14\\
 \hline\hline
 %\multirow{2}{*}{VGG16} &Single-target& \bf{99.45}$^{*}$ & \bf{91.38} & 89.49& 92.01  \\
 %&Multi-target & 98.16$^{*}$ & 90.44 & \bf{89.72} & \bf{93.32}  \\
 %\hline
 \multirow{2}{*}{Res32} &Single-target & 58.71 & 75.69 & \bf{99.70}$^{*}$ & 90.51 \\
 &Multi-target& \bf{67.69} & \bf{82.19} & 99.16$^{*}$ & \bf{93.37}  \\

 \hline
 \end{tabular}
 \end{center}
 \vspace{-1mm}
 \caption{The success rate (\%) of MANr based Single-target and Multi-target attack model. Multi-target model attack the same target as single ones.}
 \label{tab:transfer}
 \vspace{-2mm}
 \end{table}
 %-------------------------------------------------------
 In this part, we make further exploration about the transferability of single-target attack model and multi-target attack model.
 Different from last section which we take more iterations to train the multi-target attack model, 
 in this part we use same training iterations for all models.
%
% %
 For single-target models, we fix the attack target label during training,
% % 
 while multi-target models accept random input labels. 
% %
 In testing phase, we use the same attack target for multi-target models as used in single target ones.
 %, adversarial samples are generated by images in CIFAR10 test set.
% %

The results are listed in Table \ref{tab:transfer}.
% 
% According to the results, the multi-target model also improve the transferability when the single-target model performs weakly, 
According to the results, even we train two models with the same iterations, the multi-target model performs better transferability than the single-target model in all black-box cases, 
\eg it obtains 8.98\% gain when transferring from ResNet32 to VGG16.
% %
%The result further proves our conjecture that the competition between different promotes the model to learn more generalization feature with more generalization ability.
We conjecture that the competition between different target label promote the model to learn more generalization feature with more generalization ability and this result proves our conjecture.
% %--------------------------------------------------------------
% \end{figure}
 \begin{table}
 \setlength{\tabcolsep}{1mm}
 \small
 \begin{center}
 \begin{tabular}{c|c|c|c|c|c}
 \hline
  & & VGG16 & VGG19 & Res32& Res14\\
 \hline\hline
 \multirow{2}{*}{VGG16} &MANc & 97.86$^{*}$ & 87.67 & 88.26 & 88.63  \\
 &MANr & 98.55$^{*}$ & 91.07 & 88.95 & 90.33 \\
 \hline
 \multirow{2}{*}{Res32} &MANc & 51.27 & 68.11 & 98.86$^{*}$ & 81.01 \\
 &MANr& 58.56 & 72.26 & 99.26$^{*}$ & 79.33  \\
 \hline
 \multirow{2}{*}{VGG16+Res32} &MANc & 98.90$^{*}$ & 96.75 & 99.39$^{*}$ & 96.79  \\
 &MANr& \bf{99.60}$^{*}$ & \bf{98.07} & \bf{99.69}$^{*}$ & \bf{98.06} \\
 \hline
 \end{tabular}
 \end{center}
 \vspace{-1mm}
 \caption{The success rate (\%) of multi-target adversarial attack against different models on the CIFAR10 dataset. The last row indicates combine VGG16 and ResNet32 as white-box model.}
 \label{ensemble-cifar10}
 \vspace{-4mm}
 \end{table}

 \textbf{Model Ensemble.}
 We also try to combine different attacked models during the training process, the last row of Table \ref{ensemble-cifar10} shows the success rate jointly trained on VGG16 and ResNet32 models.
% %
 It is obvious that ensemble different attacked models make the transferability much better than a single architecture.
% %
 Considering the performance transferring form ResNet32 to VGG19, ensemble method exceeds the single architecture over 10\%.
% %
 This is reasonable since the network attempt to fit various architectures, thus it can be generalized well to other unseen models.

%-------------------------------------------------------------------------
\section{Conclusion}
In this paper, we propose a novel adversarial model named Multi-target Adversarial Network (MAN) to deal with the multi-target attack problem. It can produce multi-target adversarial samples by training a single model and significantly reduces the training cost and model storage. By embedding the target label information into the generated adversarial samples, a great improvement on attack ability and transferability is shown by these samples. MAN also produces diverse adversarial samples efficiently for adversarial training, which greatly improves the robustness of models.
Future work may lie on adding random noise or introducing additional regularization term to generate adversarial samples with better transferability to black-box attack. More applications about the adversarial samples are also considered to be developed, \eg adversarial attack under more realistic constraints.

\section*{Acknowledgement}
This work is supported in part by SenseTime Group Limited, in part by the General Research Fund through the Research Grants Council of Hong Kong under Grants CUHK14202217, CUHK14203118, CUHK14205615, CUHK14207814, CUHK14213616, in part by the Natural Science Foundation of China under Grant U1636201 and 61572452, and by Anhui Initiative in Quantum Information Technologies under Grant AHY150400.
{\small
\bibliographystyle{ieee_fullname}
\bibliography{egbib}
}

\end{document}